\documentclass[pdflatex,sn-mathphys-num]{sn-jnl}%

\usepackage{graphicx}%
\usepackage{multirow}%
\usepackage{amsmath,amssymb,amsfonts}%
\usepackage{amsthm}%
\usepackage{mathrsfs}%
\usepackage[title]{appendix}%
\usepackage{xcolor}%
\usepackage{textcomp}%
\usepackage{manyfoot}%
\usepackage{booktabs}%
\usepackage{algorithm}%
\usepackage{algorithmicx}%
\usepackage{algpseudocode}%
\usepackage{listings}%
\usepackage{siunitx}

\theoremstyle{thmstyleone}%

\theoremstyle{thmstyletwo}%

\theoremstyle{thmstylethree}%

\begin{document}

\title[Article Title]{An Autonomous Sea Turtle Robot that Follows Marine Animals
}

\author*[1,2]{\fnm{Zach J.} \sur{Patterson}}\email{zpatt@case.edu}\equalcont{These authors contributed equally to this work.}

\author[2]{\fnm{Emily} \sur{Sologuren}}\email{esolo@mit.edu}\equalcont{These authors contributed equally to this work.}

\author[2,3]{\fnm{Levi} \sur{Cai}}\email{cail@mit.edu}

\author[2]{\fnm{Daniel} \sur{Kim}}\email{dkim6@exeter.edu}

\author[4, 2]{\fnm{Alaa} \sur{Maalouf}}\email{alaamaalouf@cs.haifa.ac.il}

\author[2]{\fnm{Pascal} \sur{Spino}}\email{spino@mit.edu}

\author*[2]{\fnm{Daniela} \sur{Rus}}\email{rus@csail.mit.edu}

\affil*[1]{\orgdiv{Mechanical and Aerospace Engineering}, \orgname{Case Western Reserve University}, \orgaddress{\city{Cleveland}, \postcode{44106}, \country{USA}}}

\affil[2]{\orgdiv{Computer Science and Artificial Intelligence Laboratory}, \orgname{Massachusetts Institute of Technology}, \orgaddress{\city{Cambridge}, \postcode{02139}, \country{USA}}}

\affil[3]{\orgdiv{Applied Ocean Physics and Engineering Department}, \orgname{Woods Hole Oceanographic Institution}, \orgaddress{\city{Woods Hole}, \postcode{02543}, \country{USA}}}

\affil[4]{\orgdiv{Computer Science}, \orgname{University of Haifa}, \orgaddress{\city{Haifa}, \postcode{3498838}, \country{Israel}}}

\abstract{
Autonomous robots can transform how we observe marine ecosystems, enabling vision-based focal follows (animal tracking) at scale, but close-range operation in reefs and other cluttered habitats remains difficult. Vehicles must maneuver safely near animals and fragile structures while coping with currents, variable lighting and limited sensing. Previous approaches simplify these problems by leveraging soft materials and bioinspired swimming designs, but such platforms remain limited in terms of deployable autonomy. Here we present an approach to vision-based autonomy that advances the state of the art in biomimetic marine robots and we demonstrate our approach on a sea turtle--inspired underwater robot. The robot combines robust depth--heading stabilization with obstacle avoidance and target-centric control, enabling it to track and interact with moving objects in complex terrain. We validate the approach in controlled pool experiments and in an artificial coral reef at the New England Aquarium, demonstrating stable operation and reliable tracking of fast-moving marine animals. To the best of our knowledge, this is the first integrated biomimetic robotic system, combining novel hardware, control, and field experiments, deployed to autonomously follow real marine animals. With all perception and control running onboard, the robot can hold a commanded depth, navigate around obstacles (91\% success rate in the aquarium exhibit with complex lighting and visual features including glass panels) and successfully identify and follow sea turtles fully autonomously. An additional operator-initialized tracking mode extends beyond just sea turtles, including fast-moving rays, barracuda, and human divers. Together, these results establish a practical route toward bioinspired underwater robots capable of autonomous exploration and close-range animal monitoring in sensitive ecosystems.
}


\maketitle

\section*{Main Text:}\label{sec:main}
\section{Introduction}

The ocean remains one of the least explored frontiers on Earth~\cite{ramirez2010deep,danovaro2017ecosystem}, yet it plays a central role in regulating our planet's climate, biodiversity, and life systems~\cite{hoegh2010impact,sala2021protecting}. Accessing this environment requires robots that can operate reliably in unstructured, dynamic, and fragile marine habitats. While divers and traditional underwater vehicles have enabled remarkable progress in ocean exploration \cite{cardenas2024systematic, besson2022towards, yoerger2021hybrid, mccammon2026autonomous}, they typically rely on rigid hulls, thrusters, and control architectures developed for engineered vehicles rather than for animal-like locomotion~\cite{fossen2011handbook,yuh2000design,girdhar2023curee}. Biomimetic robots offer a complementary path: by drawing inspiration from the bodies and locomotor strategies of aquatic animals, they can generate movement through coordinated oscillatory motions of fins, flippers, or flexible hydrofoils~\cite{fish2006passive,lauder2005hydrodynamics,sfakiotakis1999review,zhai2025electronics,li2023underwater}. Such systems have been recognized to have the potential to combine efficient swimming, agile maneuvering, and gentle interaction with complex underwater environments~\cite{fish2020advantages,triantafyllou1995efficient,lauder2015fish}. These qualities are especially important in habitats such as coral reefs, where vehicles must maneuver near obstacles, maintain controlled motion in currents, and operate without damaging sensitive structures \cite{liu2024maneuverable, paull2013auv}.

Despite this promise, autonomy remains a major barrier to the practical deployment of biomimetic underwater robots. Many bioinspired aquatic platforms have demonstrated impressive mechanical capabilities, including deep-sea survivability~\cite{li2021self}, adaptive shape morphing~\cite{baines2022multi}, and rapid swimming speeds~\cite{chen2022performance,vandergeest2023employing}. However, these demonstrations often emphasize locomotor design or material function rather than integrated autonomous behavior in realistic environments. In particular, operation outside the laboratory requires more than untethered swimming. The robot must sense its surroundings, regulate its own motion, avoid obstacles, and execute task-level behaviors without continuous human intervention. These requirements are difficult to satisfy in biomimetic systems because the same compliant, oscillatory, and underactuated dynamics that make them attractive can also complicate integration of autonomous functionality.

As a result, the field has not yet translated biomimetic aquatic locomotion into autonomous capability. Existing platforms are frequently tested in controlled tanks, shallow pools, or structured experimental settings~\cite{wang2020development,li2023bioinspired}. Field-deployed systems have shown the feasibility of untethered or remotely guided operation in natural marine environments~\cite{katzschmann2018exploration}, but many still rely heavily on human commands, diver supervision, or simplified behavioral objectives. The central challenge is therefore not only to build aquatic robots that resemble or move like animals, but to endow them with the perception, decision-making, and control needed to act independently in the wild. Advancing autonomous biomimetic marine robots requires integrated systems that combine animal-inspired propulsion with onboard sensing, real-time control, and robust behavior in complex natural environments.

In this work, we target a specific type of ecological survey called focal following, which is the practice of following individual animals and recording their behavior. Focal follows help establish foraging routes and territory extent \cite{forsythe1997foraging}, defense and mating behaviors \cite{hanlon2020flamboyant}, and study high-resolution behavioral patterns such as feeding or defecation rates on reefs \cite{bennice2021behavioral}. This method is logistically difficult for divers because of the length of time necessary to get quality data, animal unpredictability, the inherent dangers of diving, and other factors. Because the observer must remain near the same animal for extended periods, focal following may be particularly sensitive to disturbance from the observation platform itself \cite{lopes2019effects}; underwater vehicles can elicit behavioral responses that may bias the behaviors being measured \cite{benoit2023schrodinger,rountree2010first}. Existing evidence suggests that biomimetic Autonomous Underwater Vehicles (AUVs) may therefore be especially valuable for this task \cite{kruusmaa2020salmon}, as undulatory propulsion and animal-like motion offer the potential for less acoustically and hydrodynamically disruptive observation than conventional thruster-driven vehicles~\cite{marcin2020fish,katzschmann2018exploration,pham2026characterizing}. Other than ecological relevance, focal following presents a tractable problem to address with a vision-driven robot system, and provides a useful specific benchmark to assess a platform's general autonomous capabilities. While vision-driven focal follows \textit{have} been performed with conventional AUVs \cite{cai2023semisupervised}, previous biomimetic robots have achieved tracking behaviors for targets such as light sources \cite{wang2026deeptrack} or other robots \cite{berlinger2021implicit, salazar2022multirobot}. To our knowledge no biomimetic aquatic robot has demonstrated focal following capability for real animals. 

\begin{figure} %
	\centering
	\includegraphics[width=1.0\textwidth]{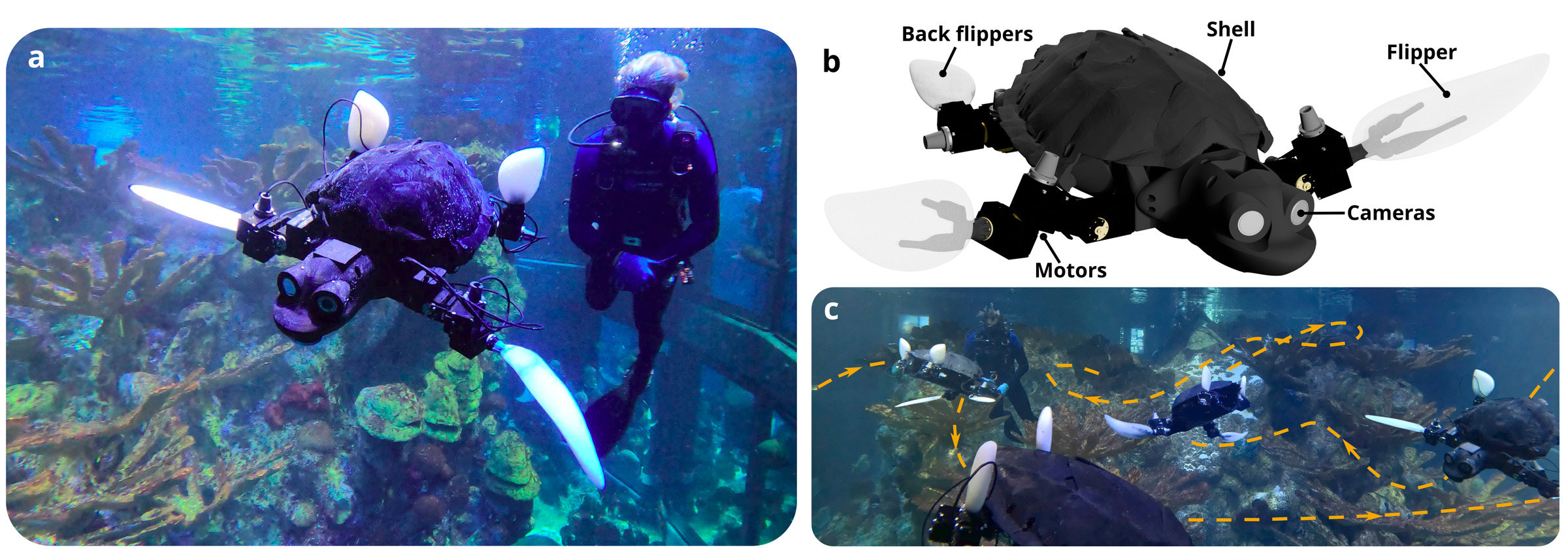}
	\caption{\textbf{Sea turtle–inspired robot for autonomous marine observation.} \textbf{a,} A photograph of the turtle robot swimming in the New England Aquarium in Boston, MA. \textbf{b,} The turtle robot uses biomimetic soft-rigid flippers and visual feedback for tracking sea animals autonomously. \textbf{c,} Superimposed frames of robot swimming autonomously and untethered around coral in aquarium tank (the orange line denotes the path of the robot).}
	\label{fig:overview} %
\end{figure}

\begin{figure} %
	\centering
	\includegraphics[width=0.9\textwidth]{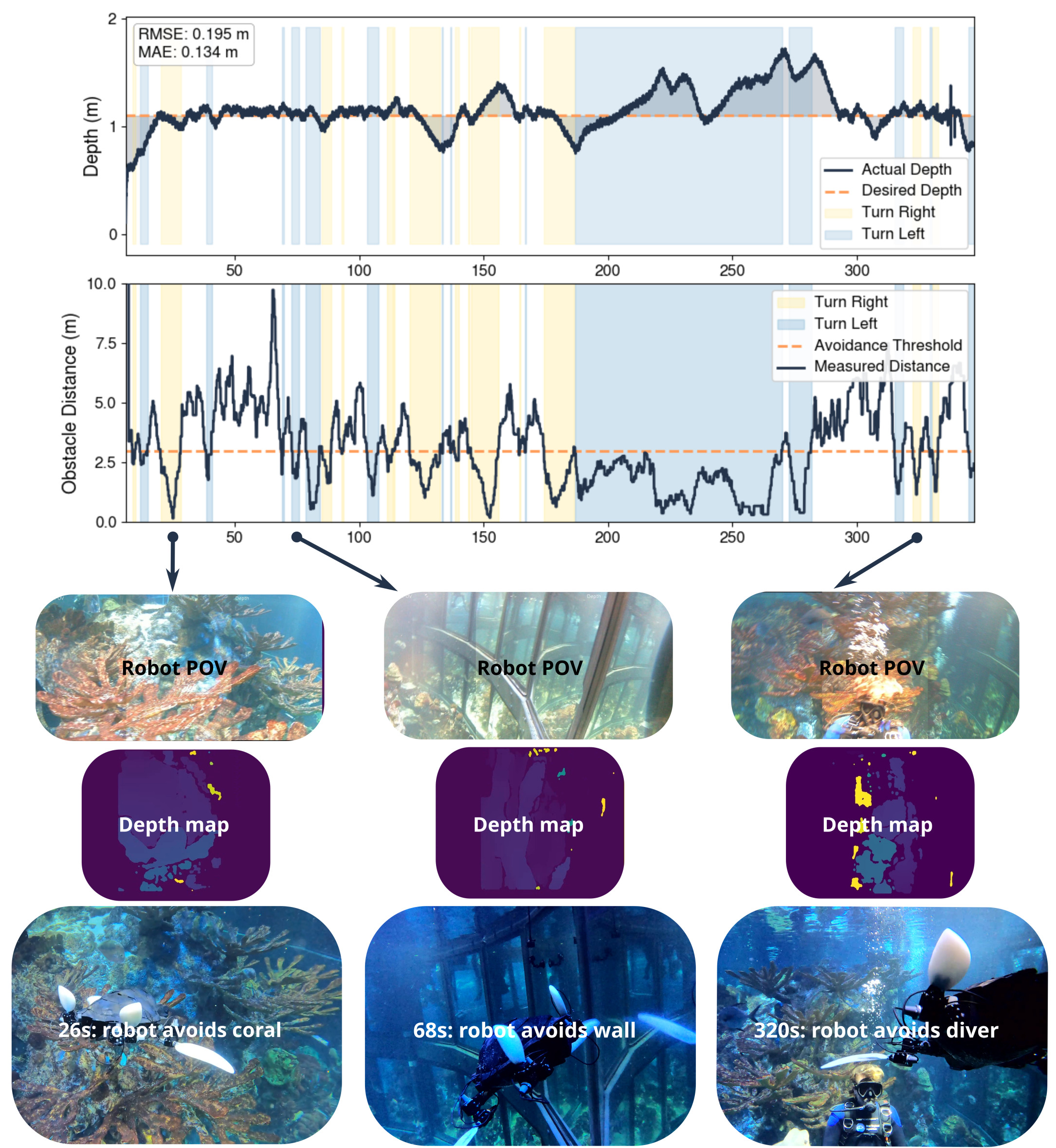}
    \caption{\textbf{3D Maneuvering around obstacles while maintaining desired depth.} \textbf{Top,} Depth over time showing the robot maintaining a target depth of 1.1\,m (RMSE\,=\,0.195\,m, MAE\,=\,0.134\,m). \textbf{Bottom,} Obstacle distance estimates over time with the 3.0\,m avoidance threshold (dashed); yellow and blue shaded regions indicate right and left turn commands, respectively. Annotated time points highlight three representative avoidance events---coral reef (26\,s), glass wall (68\,s), and a diver (320\,s)---each shown with the robot's stereo camera view (Robot POV), the corresponding stereo depth map, and an external observer perspective.}
	\label{fig:untethered} 
\end{figure}

To address this gap, we developed a vision-based system for biomimetic AUV autonomy that is demonstrated in first-of-their-kind experiments in an artificial coral reef environment around live animals. To embody our approach on a physical platform, we created Crush, a sea turtle-inspired underwater robot designed to be a testbed for agile autonomous operation in complex environments (Fig.~\ref{fig:overview}). For a video overview of our approach and results, please see Supplementary Movie 1. The testbed's design is intended to strike a balance between explicit biomimicry and pragmatic engineering necessity. The physical structure mimics the morphology of a juvenile hawksbill sea turtle (\textit{Eretmochelys imbricata}) \cite{wyneken2001anatomy}. The flippers are based on a simplified, idealized version of the skeletal and soft tissue structure of sea turtle flippers \cite{wyneken1997sea}. Skeletal components are 3D printed and overmolded into a silicone rubber matrix. The large front flippers are the main source of propulsion, while the smaller rear flippers serve as rudders. The front flippers are driven by three servomotors at each shoulder, allowing roll-pitch-yaw kinematics that are necessary for approximating sea turtle underwater flight \cite{vandergeest2022new}. The rear limbs have two degrees of freedom each, pitch and yaw. The body is printed out of rigid ABS plastic around a sealed acrylic tube hull, and the shell is directly adapted from 3D scans of a hawksbill turtle \cite{RISDNatureLab_2021_hawksbill_shell}.

The robot contains two forward-facing cameras for vision, a small downward-facing sonar module, a depth sensor, and a nine-axis inertial measurement unit (IMU). Together, these sensors enable the autonomous behaviors described below. The robot also has a detachable tether for remote monitoring, teleoperation, and utilization of off-board compute resources. A high-level system diagram is shown in Fig.~\ref{fig:system}. Crush combines several autonomous capabilities that enable collision-free swimming in cluttered environments, stable tracking of commanded headings and depths, and visual tracking of marine animals, objects, and humans. It can run fully untethered or be operated with a tether for real-time data streaming and high-level supervision, giving field teams flexibility across experimental and deployment scenarios.

A key element of the autonomy stack is a visual tracking pipeline that supports real-time following of arbitrary objects and animals. We present two tracking modalities. The primary modality runs entirely onboard, relying on the lightweight FOMO (Faster Objects, More Objects) model~\cite{edgeimpulse2022fomo} trained on a dataset of sea turtles in the New England Aquarium. It runs a standard image classification on every frame and outputs a centroid of a detected sea turtle. We also include an additional modality to support following of arbitrary objects and animals by enabling a human operator to initialize targets in the latest frame by supplying point prompts to the Segment Anything Model (SAM)~\cite{Kirillov_2023_ICCV}. The model uses the point prompt to obtain a binary mask, then tracks across subsequent frames using the CUTIE tracker~\cite{Cheng_2024_CVPR}. A lightweight health monitor detects drift and enables quick, non-disruptive re-initialization. The centroid that is outputted from either model is mapped directly to control inputs that modulate the executed gait. All gaits are defined as variations of a baseline swimming pattern through a control vector
$\mathbf{u}$. To operationalize this for tracking, pitch and yaw control are modified proportionally to the position of the tracked object with respect to the center of the frame, resulting in visual servoing. If the track is lost, the robot executes a recovery behavior by continuing to move in the direction implied by the object's last observed position.

\section{Results}

We conducted experiments in the MIT Alumni Pool and the Giant Ocean Tank (GOT) at the New England Aquarium (NEA) in Boston, MA. The GOT is 40 feet wide, 23 feet deep, and holds 200,000 gallons of saltwater. All aquarium experiments took place within the NEA exhibit, and the robot was deployed from the same observation deck located on one side of the tank. Unless otherwise stated, all results below were obtained with the robot fully untethered, with all perception and control running onboard and no external communication once submerged. Results from the operator-initialized tracking mode, which required a tether, are reported separately as an additional capability.

\subsection*{3-Dimensional Swimming}
Crush is capable of swimming autonomously along 3D trajectories. At its simplest, this requires the robot to swim forward, turn, and change depth. Additionally, it is useful for the robot to be able to roll to perform banking turns. For Crush, each of these maneuvers is produced by modulating a single baseline gait through the control vector $\mathbf{u} = [u_{\mathrm{freq}}, u_{\mathrm{roll}}, u_{\mathrm{pitch}}, u_{\mathrm{yaw}}]^T$ (Methods Section~\ref{sec:loco}),
where the $u_{\mathrm{freq}}$ controls gait frequency and thus forward speed, $u_{\mathrm{roll}}$ offsets the flipper angles asymmetrically to produce roll, $u_{\mathrm{pitch}}$ offsets the distal flipper joint to control the orientation of the flippers with respect to oncoming flow and redirect thrust to move the robot up or down the water column, and $u_{\mathrm{yaw}}$ decreases flipper amplitude in one direction to produce yaw. This control parametrization allows simple, intuitive, and continuously regulated 3D swimming. We characterized each primitive over five trials in a swimming pool (SI Section~\ref{sec:characterization}, Fig.~\ref{fig:characterization}). The robot reaches a maximum forward speed of \SI{0.32}{BL/s} (\SI{0.208}{m/s}) and a maximum yaw rate of \SI{30}{deg/s}. Its turn radius is \SI{0.29}{m} at full forward speed and shrinks to essentially zero at zero forward speed, allowing in-place rotation. It dives at up to \SI{0.18}{BL/s} (\SI{0.12}{m/s}), with repeatable descent and ascent across trials. These motion primitives enable 3D locomotion and downstream autonomous behaviors including orientation tracking, depth tracking, obstacle avoidance, and visual tracking. The robot has a total mass of \SI{10}{kg} and a dimensionless cost of transport at top speed of approximately $2.9$ ($\mathrm{CoT} = P/(mgU)$).

\subsection*{Obstacle Avoidance}

Autonomous swimming in the presence of obstacles that are natural to underwater environments, such as coral reefs, is challenging for several reasons. First, the robot cannot brake instantly, since its dynamics are underactuated, momentum-driven, and subject to currents and waves. Additionally, perception is challenging due to color attenuation and turbidity in water, and position estimates drift without an absolute reference point.

We build on a long tradition of reactive obstacle avoidance in robotics \cite{shao2005obstacle,hrabar20083d,sabe2004obstacle,huntsberger2011stereo} for the sea turtle robot. Unlike body-undulating swimmers or ground robots, a swimming sea turtle robot decouples thrust from body bending, since the body stays rigid while the limbs produce force. Thrust is generated as discrete flipper strokes with intervening glides, and control is intermittent. The obstacle avoidance decision is taken at stroke onset and must account for its effect during the glide, and the robot's heading change accumulates in discrete increments determined by the strokes. Avoiding obstacles therefore requires controlling the stroke of each flipper once per gait cycle, which differs from the traditional setting of reactive control.

To turn, the robot decreases the amplitude of one flipper relative to the other, and the yaw moment acts while the flippers are loaded. This means each stroke contributes a discrete increment of heading change, and the size of that increment is set by the amplitude difference chosen at the start of the stroke. Thus, obstacle avoidance requires many interacting decisions including when obstacles should be checked, how to predict the obstacle at the next step, and how to control the two flippers.

  Our approach addresses these with stereo-based reactive avoidance adapted to stroke-based propulsion. The robot estimates the proximity of nearby objects from onboard stereo vision, and when an obstacle's estimated distance drops below a pre-specified threshold, the robot commits to a maximum-amplitude asymmetric stroke away from the obstacle while banking with maximum roll, holding the maneuver until the obstacle clears (Methods Section~\ref{sec:untethered}). For each trial, the robot is initialized with a target depth and a nominal heading toward coral structures near the center of the tank, then allowed to operate independently for extended periods. Each obstacle encounter updates the desired heading, yielding a random-walk--like exploratory trajectory around the exhibit. There is no human intervention in guidance or control, but Aquarium divers are present to passively monitor animal safety or prevent unnecessary damage to the robot or exhibit (Fig.~\ref{fig:untethered}).

We conducted three trials in the Aquarium of durations 132, 94, and 341 seconds (mean: 189~s, standard deviation 133~s), with 10, 7, and 28 avoidance events per trial, respectively. The robot successfully avoided 41 of 45 total detected obstacles autonomously (91.1\% success rate). Only one of the four failed maneuvers involved a coral structure; the remaining three failures occurred on the glass tank walls, an artifact of the aquarium environment not present in natural reef deployments. An alternative metric of seconds per intervention (analogous to the industry-standard ``miles per intervention'' for self-driving vehicles) is 113.5~s. We further evaluated autonomous obstacle avoidance in the MIT Alumni Pool through three trials totaling 20.2~minutes, with a 95.4\% success rate across 162 avoidance maneuvers (Methods Section~\ref{sec:untethered}).

\subsection*{Autonomous Tracking}
We have extended the capabilities of the sea turtle robot with the ability to visually follow a free-swimming animal, called focal following in the ecology literature. This requires the robot to keep a moving target within its field of view while the animal changes speed, direction, and depth under its own volition, and while the robot's flippers induce pitching oscillations of the camera. Our approach uses image-based visual servoing by mapping the image-plane centroid of the target directly to yaw and pitch commands that steer the robot to keep the target centered in its view (Methods Section~\ref{sec:track}). Because the animal can briefly leave the field of view or become occluded, the robot retains the target's last observed position and continues steering toward it during short detection gaps. We developed two perception front ends for this controller: an onboard detector trained on sea turtles and an operator-initialized tracker that generalizes to arbitrary targets. We evaluate tracking and obstacle avoidance as independent capabilities, characterizing each in isolation (see Discussion).

The robot ends a track when the targeted animal stops swimming, when an arbitrary depth limit is surpassed (e.g., we stop tracking if an animal swims deeper than a perceived 7~m to maintain visibility from the observation deck), or at the discretion of the divers present for monitoring the safety of the aquarium animals. Due to time constraints and safety considerations, we leaned towards starting tracks during moments when an animal was swimming away from the observation deck, as there were points in time when a sea turtle or stingray would linger by the observation deck, too close for the robot to properly carry out a track.

\subsubsection*{Autonomous Sea Turtle Tracking}
\begin{figure} %
	\centering
	\makebox[\textwidth]{\includegraphics[width=1.1\textwidth]{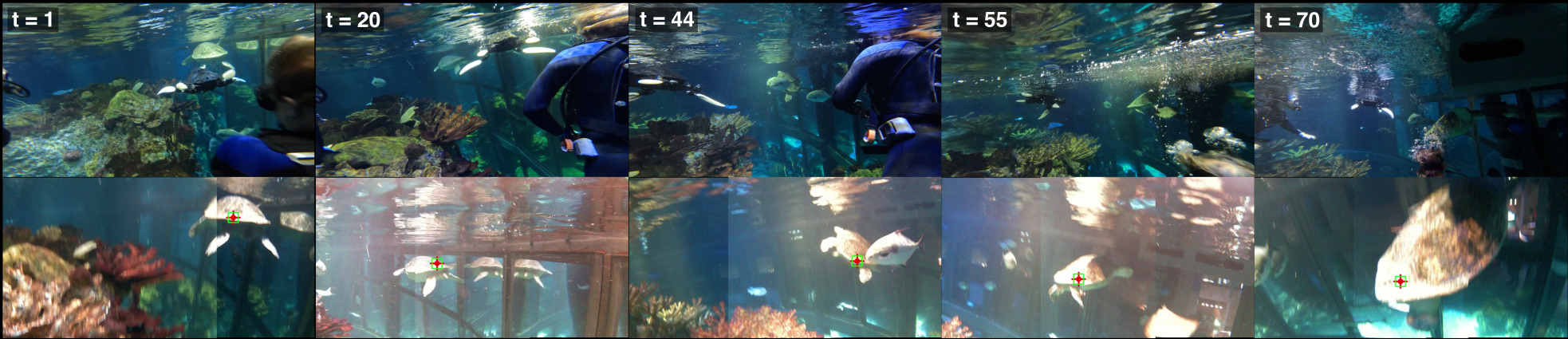}}
	\caption{\textbf{Fully onboard autonomous tracking of Sea Turtle in the New England Aquarium.} \textbf{Top,} External observer view of the robot pursuing sea turtle. \textbf{Bottom,} Onboard stereo camera view with the FOMO-detected centroid (red dot) used by the motion planner to drive the robot's motor commands in real time.}
	\label{fig:tetherless_tracking_aquarium}
\end{figure}

We deployed our tracking approach using FOMO (Faster Objects, More Objects), a lightweight object-detection model optimized for real-time inference on the robot's Raspberry Pi 5. The detector converts stitched forward-camera images into an image-plane turtle centroid that the onboard motion planner uses for closed-loop steering. Detailed model architecture, training and validation procedures, as well as target-loss handling are described in SI Section~\ref{sec:fomo_experiments}. We deploy this onboard FOMO-pipeline for live sea-turtle tracking in the same Giant Ocean Tank, the goal being to track free-swimming sea turtles Myrtle and Carolina. We carried out a total of 6 tracks with a mean tracking time of 27 seconds and a standard deviation of 20.57 seconds. Our longest track was of a 550-pound Green Sea Turtle (\textit{Chelonia mydas}) named Myrtle, which lasted 70 seconds, as seen in Fig.~\ref{fig:tetherless_tracking_aquarium}. FOMO's lightweight MobileNetV2-based architecture trades tracking robustness for onboard efficiency, and its failure modes reflect this trade-off directly. Of the three tracker-attributed failures in Table~\ref{tab:fomo_live}, one was a missed detection despite the turtle remaining in view, one a false-positive detection of coral as the target, and one a specular mirror-induced mistrack (see larger discussion on failure modes in SI Section \ref{sec:track_failures}). We attribute missed and false-positive detections to the limited size and diversity of the live-turtle training set relative to the range of orientations, lighting, and background clutter encountered in the aquarium.

Despite these detection gaps, aquarium trials, together with the model-validation results and controlled toy-target experiments (SI Section~\ref{sec:fomo_experiments}) demonstrate that the robot was able to convert onboard visual detections into closed-loop steering commands and reach a target without offboard computation. However, the current aquarium runtime data do not quantify live-animal detection accuracy, continuity during occlusion, or preservation of an individual animal's identity over time, and the onboard detector is limited to the target class on which it was trained. We next describe a complementary pipeline that extends tracking to arbitrary targets.

\subsubsection*{Generalizing to Track Anything}

A detector trained on a fixed class can follow only the animals it was trained on, yet focal follows often target species or specific individuals for which no labeled dataset exists. Following an arbitrary animal requires the robot to acquire a target specified at deployment time from a single example, then maintain that target's identity as its appearance changes with viewpoint, scale, and lighting, and through occlusions, specular reflections, and similar-looking animals nearby. We address this with a class-agnostic pipeline in which an operator selects the target with point prompts to the Segment Anything Model (SAM)~\cite{Kirillov_2023_ICCV}, and the CUTIE video object segmentation model~\cite{Cheng_2024_CVPR} propagates the target mask across subsequent frames (Methods Section~\ref{sec:track}). The resulting centroid drives the same visual-servoing controller used for onboard tracking. Because these models exceed the compute budget of the robot's onboard computer, we evaluate this pipeline as an additional capability in a tethered mode, in which the robot streams camera frames to an offboard GPU and an operator can initialize targets, monitor, and intervene. The tether serves only to provide access to offboard compute and is not fundamental to the approach; a redesign of the onboard electronics around a more capable edge computer is underway to enable this pipeline to run onboard (see Discussion). Thus, while this feature requires manual selection of the object or animal to track, it would be relatively straightforward to implement an object recognition layer to autonomously set tracks. In the meantime, this mode serves as an option for high-value deployments, such as observing a rare animal, where there are large benefits from interactivity with the operator. Regardless, these experiments can be viewed as evaluating frontier capability in open-set marine robot animal tracking. 

Each trial starts with the robot at the surface, and an operator uses a remote controller to guide the robot's heading towards a potential animal of interest. During operation, the robot sends camera frames of its point of view in real-time back to an interactive window on the laptop, allowing an operator to choose an animal to track. Once an animal of interest is in clear view on the display, the operator double clicks on the animal. This initializes the CUTIE tracker, and the robot switches into 'autonomous tracking' mode. At this point, the robot is swimming autonomously to maintain a track of the chosen animal while also sending back camera frames for the operator to monitor and potentially intervene with their remote controller should they desire.

\begin{figure} %
	\centering
	\makebox[\textwidth]{\includegraphics[width=1.0\textwidth]{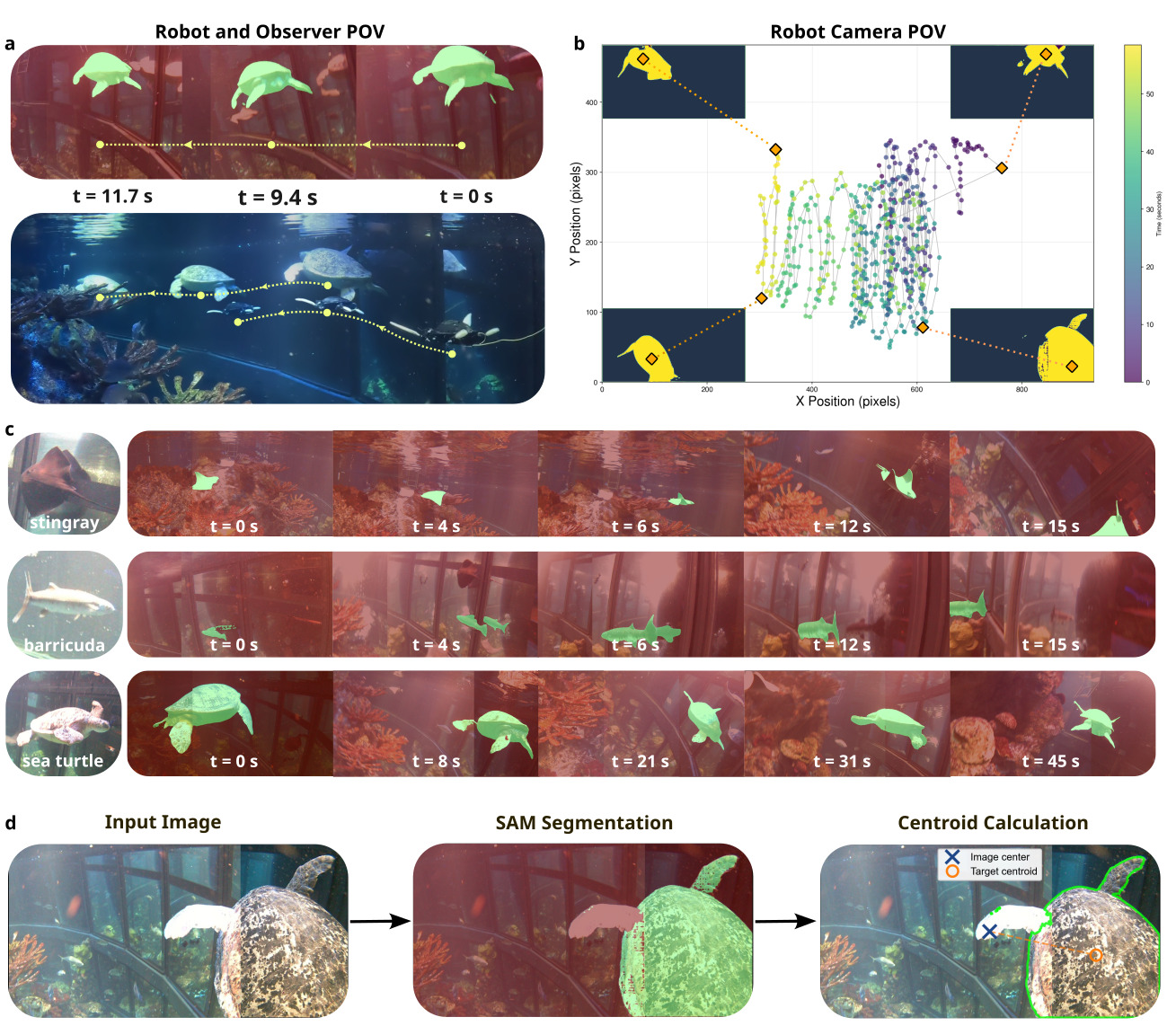}}
	\caption{\textbf{Autonomous animal tracking in the New England Aquarium.} \textbf{a,} Time series showing the robot's point-of-view (POV) with target segmentation masks (green overlays) during autonomous sea turtle tracking over $\sim$12 seconds. \textbf{b,} Target centroid trajectories throughout a minute-long tracking sequence, with corner indices marking trajectory extremities (top left, top right, bottom left, and bottom right). \textbf{c,} Additional tracking examples: stingray track ending when the animal dived beneath the robot, barracuda track maintaining target lock despite specular reflections, and sustained sea turtle pursuit. \textbf{d,} Tracking pipeline: stereo cameras are stitched into a wide-field view, segmented to extract target centroid, which drives autonomous swimming control.}
	\label{fig:tethered}
\end{figure}

Successful tethered-mode tracks include individual sea turtles, barracudas, stingrays, and divers. The longest track was of Myrtle (Fig.~\ref{fig:tethered}), following her from one end of the tank to the other. As seen in Fig.~\ref{fig:tethered}B, the tracker generally maintained its track of Myrtle towards the center portion of the robot's point of view. It can be seen that when the track was first initialized (at t=0 in A of \ref{fig:tethered}), the centroid started towards the top right of the robot's point of view, with the centroid eventually moving towards the center. By the end of the track, we note that the recorded centroids leaned towards the left side of the robot, as this was when the sea turtle stops swimming. During the whole track, we observe that the centroids over time move in an oscillatory pattern due to the pitching motion of the robot. Despite this, we maintained either a full or partial view of Myrtle throughout the duration of the track. We conducted a total of 20 sea animal tracks, with a mean track time of 25.85 seconds and a standard deviation of 14.7 seconds. Of those 20 tracks, 8 were green sea turtles (\textit{Chelonia mydas}), 4 loggerhead sea turtles (\textit{Caretta caretta}), 6 were cownose rays (\textit{Rhinoptera bonasus}), and 2 were great barracuda (\textit{Sphyraena barracuda}). Most tracks ended because the animal stopped moving, or because the robot was entering a portion of the tank that the NEA team preferred to avoid. A few tracks, particularly of the rays, were difficult due to the speed discrepancy between the animal and the robot. Cownose rays are reported to have a typical cruising speed of about \SI{1}{m/s} \cite{hao2025equivalent}, significantly faster than the robot. A good example of this can be seen in C of \ref{fig:tethered}, where the robot maintains a track of a cownose ray for 12 seconds and ends when the stingray both speeds up and changes direction. Our longest tracks were about a minute, which ended because the animal stopped swimming (Table~\ref{tab:track_analysis}, full results in Table~\ref{tab:track_full_analysis}). To assess the capabilities of the tracker in a controlled environment, we collected over 45 minutes of tracking several inanimate objects and a human swimmer at the MIT Alumni Pool. We performed these tracks for periods up to 249 seconds, and the tracker was highly robust to failure, with no failed tracks. However, there were instances of failure at the aquarium---four times due to specular/mirror-induced duplicates, low-light/low-contrast regimes, and stitching line distortions. 

\begin{table}[ht]
\centering
\caption{\textbf{Summary of notable operator-initialized aquarium tracking experiments.}}
\label{tab:track_analysis}
\begin{tabular}{@{}lccc@{}}
\toprule
Animal type & Track time (s) & Failed & End track reason \\
\midrule
Green sea turtle (Myrtle) & 59 & No & Stopped swimming\textsuperscript{a}\\
Green sea turtle (Myrtle) & 57 & No & Stopped swimming\textsuperscript{a}\\
Green sea turtle (Myrtle) & 43 & No & Visibility\\
Green sea turtle (Myrtle) & 38 & No & Coral collision\textsuperscript{b}\\
Cownose ray & 32 & No & Swam over robot \\
Cownose ray & 30 & No & Too fast \\
Cownose ray & 14 & Yes & Tracker failure \\
Great barracuda & 20 & No & Depth limitation \\
Great barracuda & 15 & Yes & Tracker failure \\
\bottomrule
\end{tabular}
\vspace{0.3cm}
\begin{flushleft}
\footnotesize
Several aquatic animals were observed and tracked in the Giant Ocean Tank. Full results including pool trials are provided in Table~\ref{tab:track_full_analysis}. Green sea turtles are \textit{Chelonia mydas}; cownose rays are \textit{Rhinoptera bonasus}; great barracuda are \textit{Sphyraena barracuda}. Only the aquarium's named individuals, Myrtle and Carolina (seen in Table~\ref{tab:track_full_analysis}), are identified individually; other individuals within each species were not distinguished across tracks.

\textsuperscript{a} some tracks end early for the safety of the aquarium animals \\
\textsuperscript{b} track was not lost but ended early to prevent damage to robot
\end{flushleft}
\end{table}

\section{Discussion}

To our knowledge, these are the first demonstrations of a biomimetic robot used to track and monitor real marine animals~\cite{prakash2024bioinspiration}. During tracking, we observed remarkable recovery from occlusion and objects leaving the robot's field of view. During fully autonomous untethered operation, we achieved extended deployments in environments containing challenging visual features such as glass-panel reflections in the NEA Giant Ocean Tank and variable artificial lighting throughout the exhibit.

These results on challenging, real-world tasks in complex environments establish a baseline and open several promising directions. On the perception side, our full track-anything pipeline currently exceeds the compute budget of the low-cost Raspberry Pi onboard the robot. A redesign of the onboard electronics around a more capable edge platform (e.g., an NVIDIA Jetson) is underway to run this pipeline without a tether. In parallel, we will further refine our lightweight embedded tracker through model optimization and temporal association, validated against frame-level ground-truth annotations of onboard trials. On the hardware and control side, co-optimizing flipper morphology and gait parameters offers a natural path to faster, more efficient swimming, enabling longer-range missions and more reliable tracking. For better field readiness, increasing battery capacity by using a larger enclosure would also directly enable greater endurance and adding lights would enable low-light missions. A further milestone toward field-ready autonomy is combining the independently validated obstacle-avoidance and tracking behaviors into a single closed-loop mission. For example, we may arbitrate between avoidance and pursuit when they issue conflicting steering commands. For extended autonomous deployments, navigation is another area to explore. Most conventional AUVs rely on Doppler Velocity Loggers (DVLs) for dead reckoning, and we plan to integrate compact DVLs~\cite{hasan2024oceanic} alongside long-range navigation strategies tailored to biomimetic AUVs~\cite{gill2024navigation}. More structured recovery behaviors---both for re-acquiring lost targets and for disentangling from complex environments---could further enhance reliability and mission duration.

Recent studies have suggested that biomimetic robots can reduce avoidance responses in wild animals, and that turtle-like forms in particular reduce flee responses in many organisms~\cite{asunsolo2023behaviour,kruusmaa2020salmon}. In our experiments, the animals tracked in the aquarium showed no noticeable change in behavior with the introduction of Crush, other than one turtle approaching and biting the yellow tether in early experiments. Recent field work with our Crush robot have shown that the platform has potential to be less disturbing than a conventional AUV \cite{pham2026characterizing}. However, further studies are needed to understand animal-robot interactions in the wild and to validate the potential value of biomimetic robots for marine observational science \cite{jessop2024comparison, cai2025measuring}.

\section{Methods}\label{methods}

\subsection{Fabrication and Hardware}
Crush houses all its electronics in a \SI{100}{mm} inner diameter, \SI{300}{mm} long Blue Robotics watertight enclosure tube, made from cast acrylic plastic and rated to depths up to \SI{1000}{m} in salt water. Its onboard electronics include a Raspberry Pi 5, motor controllers, a pressure sensor, 9-axis IMU, and various components for communication, power management, and heat dissipation. The battery is a 3S 8000~mAh battery. For vision, we embed two exploreHD 3.0 (\SI{400}{m} depth rated) Underwater ROV/AUV USB cameras into the turtle head. The flippers are injection molded using a custom molding setup. Each flipper consists of a loosely bioinspired 3D-printed skeleton, which constrains the motion of the proximal flipper, and a silicone matrix overmolded onto the skeleton to provide flexible hydrofoil dynamics~\cite{patterson2025design,desatnik2023soft,bell2022injection}. The flippers are attached to 3-DOF shoulders consisting of Dynamixel XW540-T260-R motors. To insulate the connector-to-motor interface from saltwater, we applied marine epoxy; after this modification, the motors reliably functioned for hour-long deployments at approximately \SI{10}{m} depths. The turtle's head, chassis, and shell were printed from ABS on a Bambu P1S with 100\% infill to prevent water absorption.

\subsection{Biomimetic locomotion}\label{sec:loco}
Crush's locomotion policy is based on trajectories derived from motion capture analysis of green sea turtles characterized by Van der Geest et al.~\cite{vandergeest2022new}. These sinusoidal trajectories specify a straight swimming gait for 3-DOF shoulder joints, and are time-parametrized so that speed can be altered by adjusting frequency. To achieve pitch, yaw, and roll control, we apply a series of transformations to the base gait. For pitch control, we offset the distal flipper joint, altering the angle of attack and the direction of the resulting thrust vector:
\begin{equation}
\mathbf{q}_{\mathrm{des, pitch}} = \mathbf{q}_{0,\mathrm{pitch}} + \theta_{\mathrm{max}} \, u_{\mathrm{pitch}},
\label{eq:pitch_control}
\end{equation}
where $\mathbf{q}_{0,\mathrm{pitch}}$ is the baseline gait, $\theta_{\mathrm{max}}$ is the maximum pitching angle, and $u_{\mathrm{pitch}} \in [-1,1]$. For yaw control, we reduce the amplitude of the limb movement on the turning side:
\begin{equation}
\mathbf{q}_{\mathrm{des}} = \mathbf{q}_{0} \cdot (u_{\mathrm{pitch}} + 1.0).
\label{eq:yaw_ctrl}
\end{equation}
Roll control applies the same transformation as pitch but with opposing hydrofoil directions to generate a rolling torque. These trajectories are executed via a joint-space PD controller:
\begin{equation}
\boldsymbol{\tau}(t) = \mathbf{K}_p\bigl(\mathbf{q}_{\mathrm{des}}(t) - \mathbf{q}(t)\bigr) + \mathbf{K}_d\bigl(\dot{\mathbf{q}}_{\mathrm{des}}(t) - \dot{\mathbf{q}}(t)\bigr),
\label{eq:flipper_control}
\end{equation}
where $\mathbf{K}_p$ is a diagonal proportional gain matrix tuned for our platform~\cite{lynch2017modern}. The rear flippers act as rudders for pitch stabilization, oscillating at the front flipper gait frequency to counteract pitching oscillations of the head and camera. Although not biomimetic, we found this feature to be practically useful for camera-driven autonomy in lieu of a neck.

\subsection{Track Anything pipeline}\label{sec:track}
We implement a real-time, human-in-the-loop tracking pipeline composed of two phases: (1) interactive object initialization via point-prompted segmentation and (2) online mask propagation. Initialization is performed with SAM~\cite{Kirillov_2023_ICCV} and propagation with the CUTIE tracker~\cite{Cheng_2024_CVPR}.

A dedicated capture thread continuously reads frames from the live stream and maintains a single-element buffer, discarding stale frames to minimize latency. A rendering thread overlays the current segmentation mask and centroid. To initialize a target, the operator provides point prompts: positive prompts (left-clicks) placed inside the object of interest, and negative prompts (right-clicks) placed outside. Upon receiving three total prompts within $\Delta t \le 3$~s (with at least one positive), the system invokes SAM on the latest captured frame. For each subsequent frame, the current binary mask and next frame are passed to CUTIE, which predicts the next-frame mask and outputs the centroid to the controller.

Lightweight health signals (mask area change, centroid displacement, and tracker confidence) detect anomalies. A two-right-click gesture within 2~s triggers corrective re-initialization: the operator supplies new point prompts, SAM is re-invoked, and the tracker is warm-restarted. The system is formalized as a non-blocking state machine with states \textsc{Idle}, \textsc{Init}, \textsc{Track}, and \textsc{Correct}.

Given a tracked centroid position $[x, y]$ in a frame with width $w$ and height $h$, visual servoing control inputs are:
\begin{gather}
u_{\mathrm{yaw}} = 2x/w - 1, \\
u_{\mathrm{pitch}} = 2y/h - 1.
\end{gather}
If the track is lost, the robot recovers by moving in the direction implied by the object's last successfully tracked position.

Pool tracking experiments used a variety of objects: toy turtle, toy pig, human head, fish tail, and a free-swimming human subject. In total, we acquired approximately 45 minutes of tracking time with minimal tracking errors.

\subsection{Untethered Autonomy}\label{sec:untethered}
\subsubsection{Obstacle Avoidance}
A small USB-powered WiFi router is used to initialize the robot before submersion; once submerged, no external communication is possible. Crush's two forward-facing stereo cameras (\SI{54}{mm} baseline, 640\,$\times$\,480~pixels, \SI{30}{Hz}) are calibrated using the fisheye camera model in OpenCV~\cite{opencv_library} via checkerboard calibration performed underwater.

Obstacle proximity estimation proceeds as follows. Raw left and right images are undistorted and rectified using precomputed mapping functions, then stereo block matching (StereoBM) computes the disparity map. The raw disparity map is filtered to smooth depth discontinuities and suppress invalid disparities. The filtered disparity is converted to a highly approximate version of metric depth using the reprojection matrix $\mathbf{Q}$:
\begin{equation}
z(u,v) = \frac{Q_{2,3}}{Q_{3,2} \cdot d(u,v)},
\end{equation}
where $z(u,v)$ is proximity in meters at pixel $(u,v)$ and $d(u,v)$ is disparity. Pixels with proximity less than \SI{3.0}{m} are classified as obstacles and segmented via binary thresholding. OpenCV contour detection identifies contours exceeding 3\% of the field of view, with temporal filtering every 15 frames to reduce false detections. Due to the fact that the cameras are not synchronized, these distance estimates should not be considered accurate quantitative measurements, but they do provide useful course signals for obstacle avoidance.

During an obstacle encounter, the robot is commanded to produce maximum yaw ($u_{\mathrm{yaw}} = \pm 1$) opposite the centroid of the main contour, while also commanding a maximum roll ($u_{\mathrm{roll}} = \pm 1$) to bank along the obstacle. The maneuver continues until the filtered stereo depth reading rises above the threshold. All image processing runs onboard at approximately \SI{10}{Hz}.

\subsubsection{Visual Tracking (FOMO)}\label{sec:fomo_methods}

For untethered animal tracking, we replace the tethered SAM/CUTIE pipeline (Section~\ref{sec:track}) with FOMO (Faster Objects, More Objects), a lightweight centroid-detection model trained via Edge Impulse and compiled for onboard inference on the Raspberry Pi 5. Stitched stereo frames are passed through the detector, which outputs a single turtle centroid per frame or a null result when no candidate exceeds the application-level confidence threshold.

The selected checkpoint achieved centroid-assignment precision of 0.809, recall of 0.899, and F1 of 0.852 on the validation partition (SI Section~\ref{sec:fomo_experiments}); these are centroid-assignment statistics under a distance-tolerant scoring rule, not conventional bounding-box precision, recall, or mean average precision.

The motion planner consumes the detected centroid using the same visual-servoing control law as Section~\ref{sec:track}. During short detection gaps, the planner retains the last observed centroid for up to nine consecutive missed frames; after ten consecutive missed frames, the forward, yaw, and pitch tracking commands are set to zero. This short-term fallback is implemented entirely in the motion planner. The FOMO model itself does not maintain temporal target identity or associate detections across frames.

Full architecture, training procedure, dataset composition, and validation methodology are reported in SI Section~\ref{sec:fomo_experiments}.

\backmatter

\bmhead{Acknowledgements}
We would like to thank the New England Aquarium for allowing us to conduct experiments in their Giant Ocean Tank, and we would like to thank MIT DAPER for use of their swimming pool for experiments. 
\section*{Declarations}

\begin{itemize}
\item Funding: SMART M3 program, GIST Physical AI, BARI
\item Conflict of interest: There are no competing interests to declare.
\item Ethics approval: The experimental protocol at the New England Aquarium was approved by an IACUC, protocol number 2024-10.
\item Data availability: Data is available at \url{https://github.com/zpatty/drl-turtle}
\item Supplementary information: Supplementary Movie 1.
\item Materials availability: N/A
\item Code availability: Code is available at \url{https://github.com/zpatty/drl-turtle}
\item Contributions: DR contributed to conceptualization, idea formation and aims, analysis, methodology, experimentation and validation, writing, and funding. ZJP contributed to conception, design, software, experiments, and writing. ES contributed to design, software, experiments, and writing. LC contributed to experiments and writing. DK contributed to experiments and writing. AM contributed to software and writing. PS contributed to experiments and writing.
\end{itemize}

\noindent
\bibliography{bibliography}%

\newpage
\clearpage
\section*{Supplementary Information}

\newcommand{\beginsupplement}{
    \setcounter{section}{0}
    \renewcommand{\thesection}{S\arabic{section}}
    \setcounter{equation}{0}
    \renewcommand{\theequation}{S\arabic{equation}}
    \setcounter{table}{0}
    \renewcommand{\thetable}{S\arabic{table}}
    \setcounter{figure}{0}
    \renewcommand{\thefigure}{S\arabic{figure}}

    \renewcommand{\theHsection}{S\arabic{section}}
    \renewcommand{\theHequation}{S\arabic{equation}}
    \renewcommand{\theHtable}{S\arabic{table}}
    \renewcommand{\theHfigure}{S\arabic{figure}}
}

\beginsupplement

\begin{figure}[ht] %
	\centering
	\includegraphics[width=0.8\textwidth]{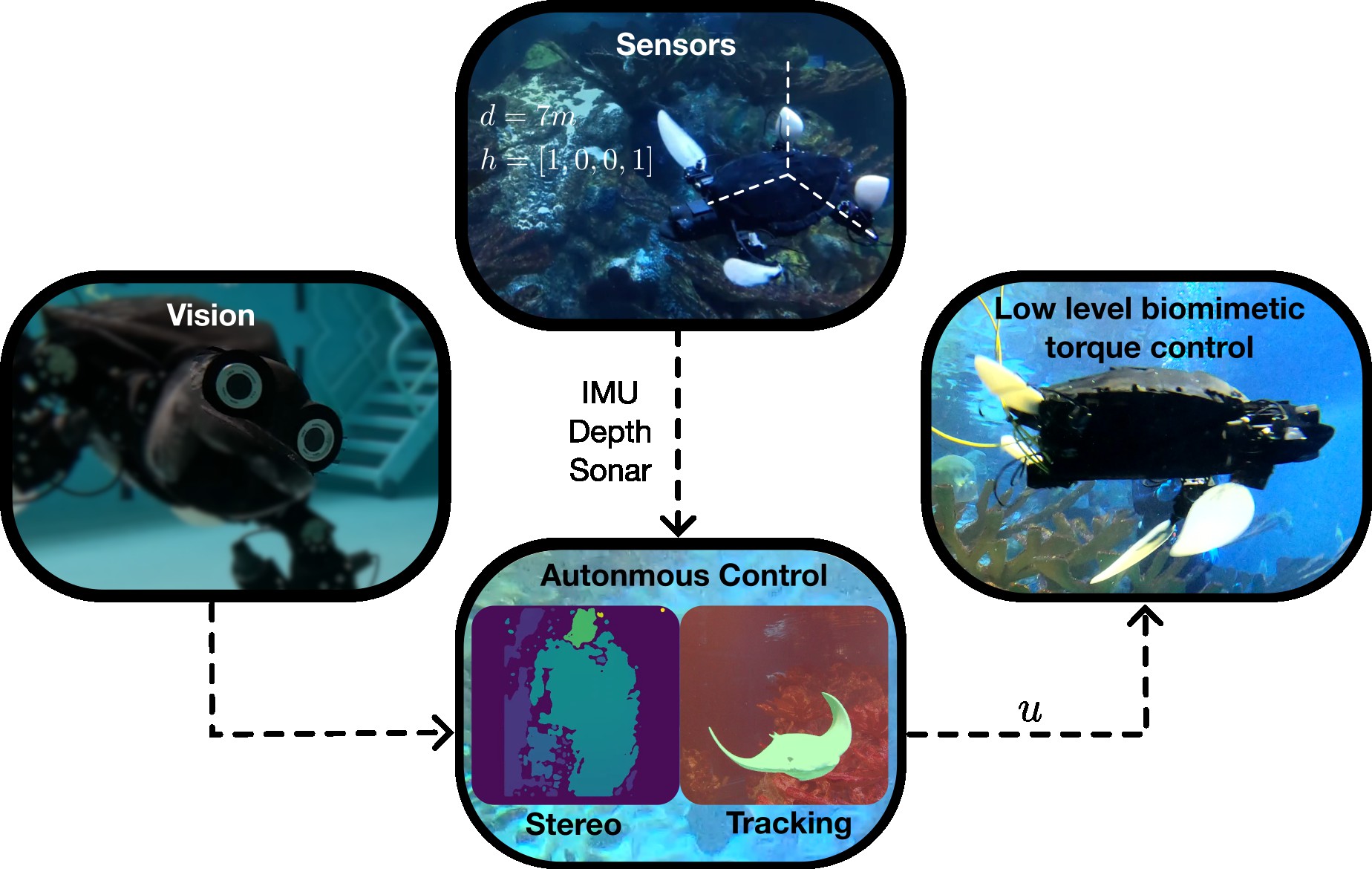}
	\caption{\textbf{System Diagram of Crush's Autonomous Control and Tasking:} the robot is equipped a suite of cameras and sensors to enable high-level vision-driven autonomy, while a lower level torque control loop handles biomimetic locomotion to produce the desired high-level motion commands.}
	\label{fig:system} %
\end{figure}

\begin{figure} %
	\centering
	\includegraphics[width=0.8\textwidth]{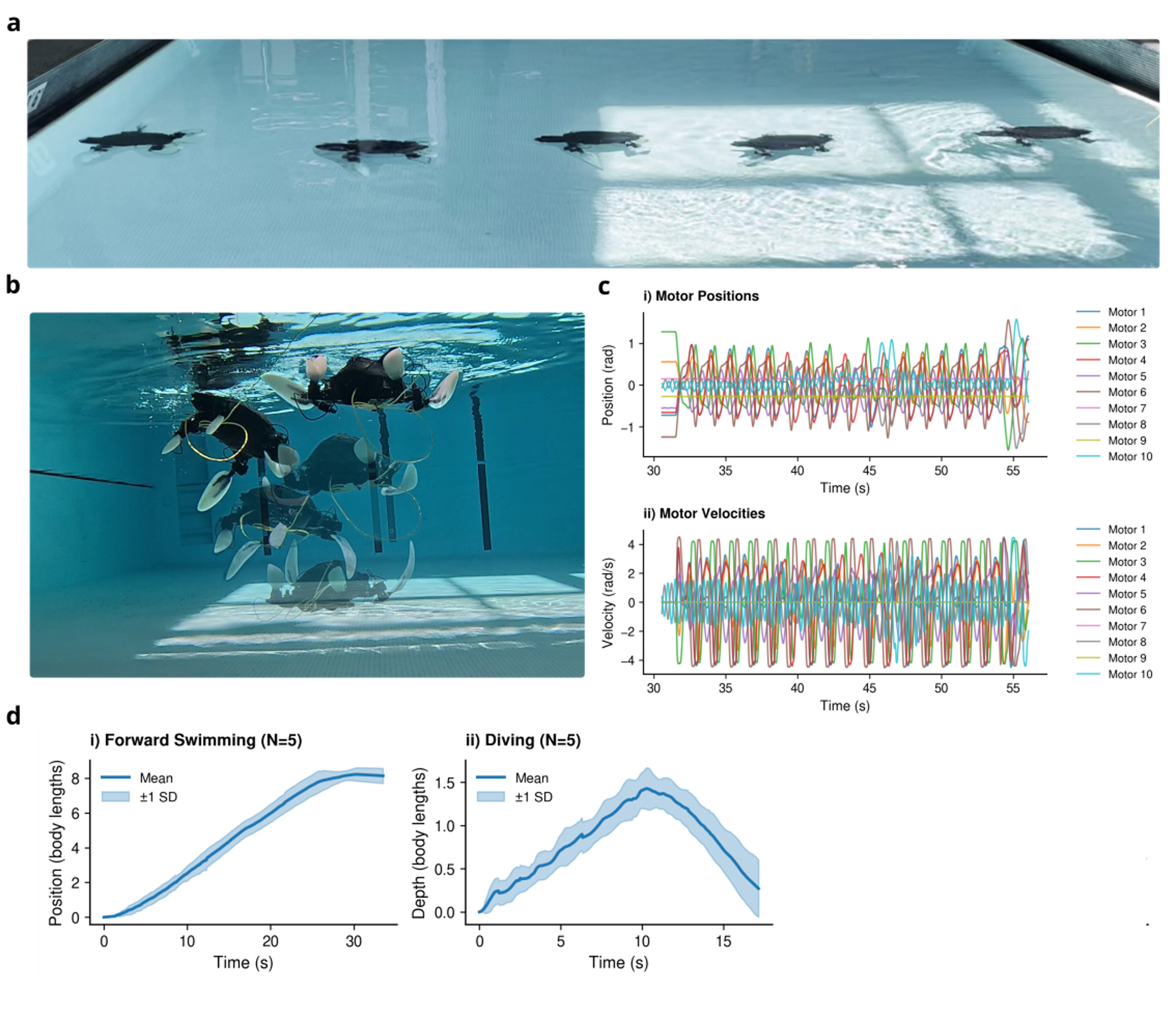}
    \caption{\textbf{Robotic Sea Turtle Characterization:} 
	\textbf{a)} Image sequence showing the robot's forward swimming trajectory in a pool environment. 
	\textbf{b)} Underwater view of the robot during a diving maneuver, demonstrating flipper-based propulsion and pitch control. 
	\textbf{c)} Motor kinematics during steady swimming: \textbf{i)} joint positions showing synchronized oscillatory flipper motion across all 10 motors (front and rear flippers), and \textbf{ii)} corresponding motor velocities demonstrating consistent stroke patterns. 
	\textbf{d)} Performance characterization across N=5 trials: \textbf{i)} Forward swimming speed showing mean position $\pm$ 1 standard deviation over 30 seconds, \textbf{ii)} diving behavior showing depth excursion of approximately 1.5 body lengths with controlled ascent and descent phases.}
	\label{fig:characterization}
\end{figure}

\section{Turtle Characterization}\label{sec:characterization}

Characterization was performed on maximum forward swimming speed, straight-line diving speed, and yaw rate, each over five trials in a swimming pool. The swimming pool's dimensions are 12.2m x 6.1 m, with its most shallow depth at 1.1m and deepest depth of 1.4m. Robot motion was tracked by external cameras; pixel distances were converted to body lengths by measuring the robot length in pixel space. Forward swimming and diving trajectories were tracked semi-automatically using the Kanade-Lucas-Tomasi (KLT) algorithm~\cite{tomasi1991detection} applied to a manually selected keypoint on the robot. Yaw rate was computed from frame counts per full rotation. Results are reported in Fig.~\ref{fig:characterization} as the average across the five trials.

The robot swims with a maximum forward speed of \SI{0.32}{BL/s} (\SI{0.208}{m/s}) and a maximum yaw rate of \SI{30}{deg/s}. At maximum forward velocity, the turn radius of the robot is \SI{0.29}{m}, while at 0 forward velocity the turn radius is near to \SI{0}{m}. The robot dives with a maximum speed of \SI{0.18}{BL/s}. The mass is \SI{10}{kg}. The cost of transport at top speed is approximately $2.9$, which we derive from an estimated \SI{1.5}{hr} endurance at this speed and a total onboard battery capacity of \SI{320}{kJ}. In endurance tests across all experiment types, we observed a maximum battery life of about 90--105 minutes. In the controlled dive experiments at the pool, we demonstrate the robot is capable of consistently diving and surfacing from a depth of \SI{1.0   }{m}. However, from the deployment experiments in the wild we found that the same robot was capable of diving to roughly \SI{10}{m} \cite{pham2026characterizing}. 

We additionally provide a comparison table that breaks down both the physical and autonomous differences between Crush and a selection of other biomimetic robots found in the literature. As seen in table \ref{tab:comparison}, we recognize that other robots have better locomotive efficiency, such as the Tunabot Flex with a cost-of-transport of 1.88. However, our robot's mass and cost-of-transport is comparable to other biomimetic robots, with the additional benefit of having onboard depth control, perception, power, and proven animal tracking capability. We hope that with future iterations we can get closer to the physical performance of Tunabot, other frontier biomimetic swimmers, and real animals. 

\begin{figure}
	\centering
	\makebox[\textwidth]{\includegraphics[width=1.1\textwidth]{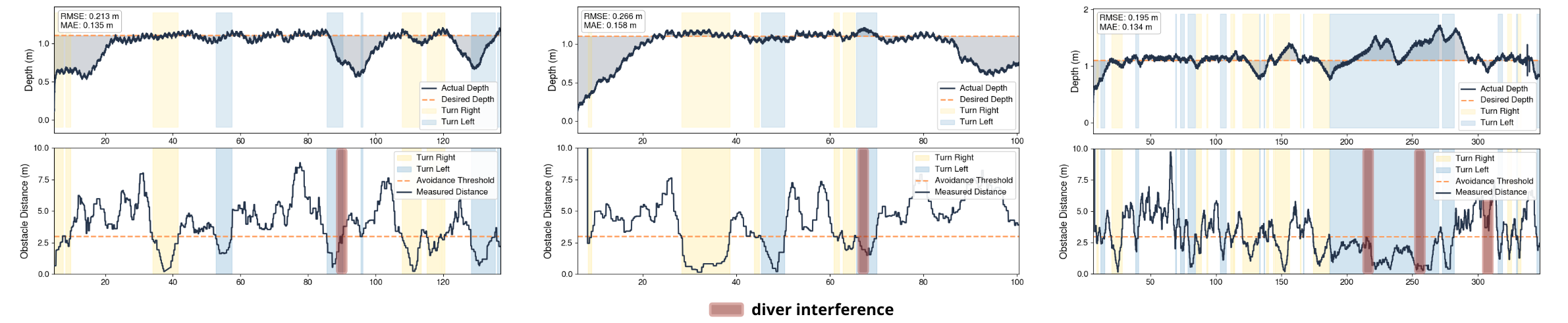}}
	\caption{\textbf{Three untethered obstacle avoidance trials at the New England Aquarium}
(132\,s, 94\,s, and 341\,s; left to right). For each trial, depth over time
(top) and estimated obstacle distance (bottom) are shown, with yellow and
blue shaded regions indicating right and left turn commands, respectively, and
red shaded regions marking diver interference events. The robot maintained a
1.1\,m target depth across all trials (RMSE\,=\,0.213, 0.266, and 0.195\,m;
MAE\,=\,0.135, 0.158, and 0.134\,m) while executing 10, 7, and 28 avoidance
maneuvers per trial, respectively. Across all three trials, the robot
autonomously avoided 41 of 45 detected obstacles (91.1\% success rate,
113.5\,s per intervention), with the three wall-collision failures attributable
to the glass tank environment rather than reef-relevant obstacles.}
	\label{fig:obstacle_avoidance_si}
\end{figure}

\section{CUTIE Tracker Failures}\label{sec:track_failures}
While the CUTIE tracker was remarkably robust, we observed three failure modes: (i) \textit{Specular/mirror-induced duplicates.} When a target is adjacent to a mirror-like surface, the specular image is visually coherent with the real object and moves in a correlated manner, leading to ID switches or mask mixing. (ii) \textit{Low-light/low-contrast regimes.} When the tracked object enters a dark or far zone, per-pixel tracking likelihoods become noise-dominated, suppressing detections and triggering track termination. (iii) \textit{Stitching line distortions.} Fast motion across or along the stitching seam between the two camera views produces seam-level splits or local distortions that break appearance/motion continuity.

\section{Untethered Tracking}\label{sec:fomo_experiments}
Untethered FOMO (Faster Objects, More Objects) tracking was evaluated in two stages using separately trained detectors. 
The motion planner's centroid-fallback logic is described in Methods Section~\ref{sec:fomo_methods}.
FOMO was used for centroid localization rather than prediction of the turtle's full spatial extent. The $30\times30$-pixel rectangles written to the runtime log and displayed around detected centroids were fixed visualization markers and were not predictions of the turtle's width or height.

An initial controlled proof-of-concept detector was trained on 2,760 images of toy turtles captured in the MIT Sea Grant tank and Alumni Pool at various distances and orientations using a 75/25 train-test split. Images containing no target were retrained as negative observations. The model was compressed to float32 binary format and integrated into the perception pipeline. Incoming stereo images were passed through the model, which outputs a centroid location when a target is detected and a null result otherwise. This detector was used for the controlled toy-target experiments described below and is distinct from the live-turtle detector subsequently developed for aquarium deployment.

\begin{figure} %
	\centering
	\includegraphics[width=0.9\textwidth]{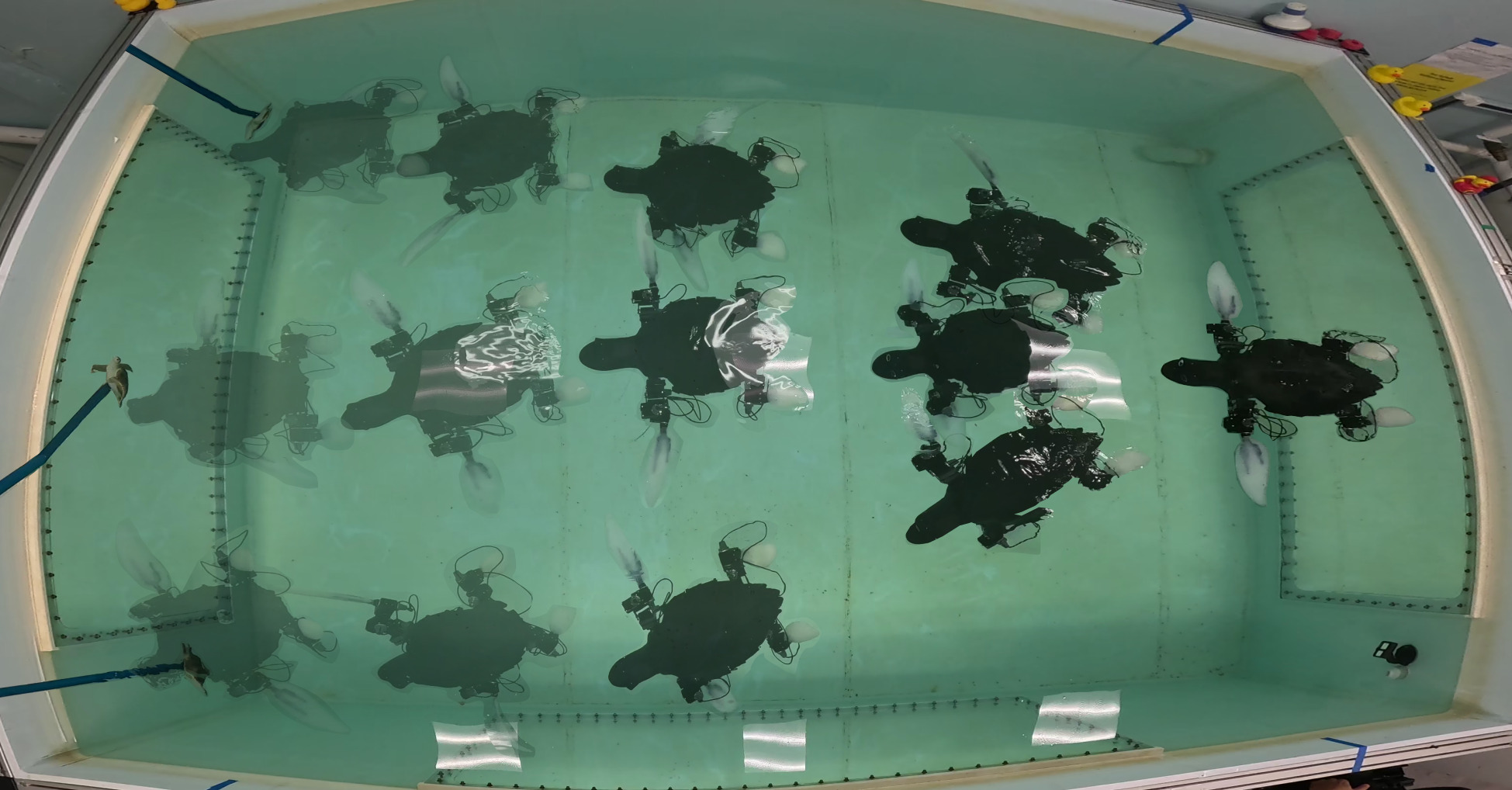}
	\caption{\textbf{Tetherless Tracking Trials.} Top-down view of the autonomous tracking robot pursuing a toy turtle target across multiple trials in the MIT Sea Grant testing tank. Each semi-transparent silhouette represents the robot at successive positions during left, right, and center target trials. The glass wall at the center of the tank presented detection challenges, requiring the target to be repositioned closer to the robot for reliable model detection.}
	\label{fig:seagrant} %
\end{figure}

\begin{figure} %
	\centering
	\makebox[\textwidth]{\includegraphics[width=1.1\textwidth]{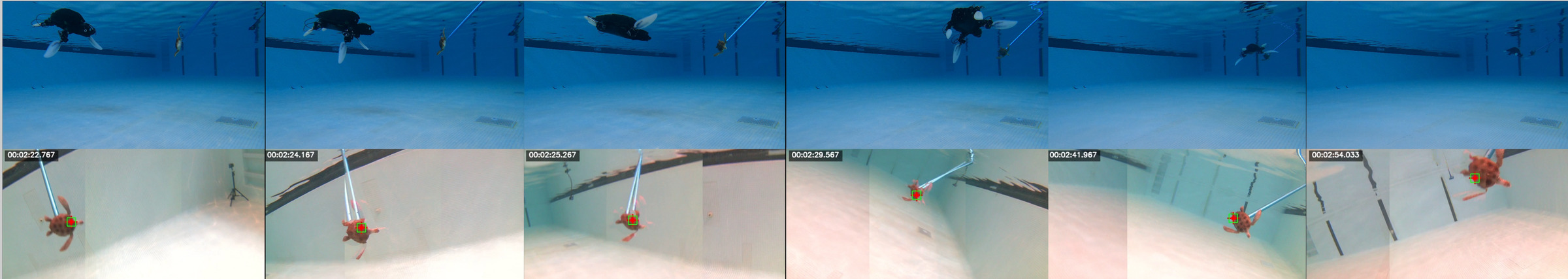}}
	\caption{\textbf{Onboard tetherless tracking in the MIT Alumni Pool.} \textbf{Top,} External observer view of the robot pursuing the toy turtle. \textbf{Bottom,} Onboard stereo camera view with the FOMO-detected centroid (red dot) used by the motion planner to drive the robot's motor commands in real time. Segment shown is 34 seconds from a 234-second autonomous tracking trial.}
	\label{fig:alumni_pool}
\end{figure}

We validated this model in the MIT Sea Grant testing tank (3.95\,m $\times$ 2.95\,m $\times$ 1.09\,m) with targets placed at leftmost, rightmost, and center positions (Fig.~\ref{fig:seagrant}). Mean time to reach the toy turtle for left and right targets was 15.83 and 15.85~s, respectively. For the center position, mean track time was 20.91~s, as the glass wall at the tank center presented detection challenges. A 34-second segment from a 234-second autonomous tracking sequence is shown in Fig.~\ref{fig:alumni_pool}; 

Following the controlled toy-target proof of concept, we developed a separately trained FOMO detector for deployment at the New England Aquarium. The architecture, training, validation, and aquarium runtime results reported below refer to this new live-turtle dataset detector unless otherwise stated.

The onboard detector used the FOMO (Faster Objects, More Objects) architecture implemented in an Edge Impulse custom learning block and trained locally using TensorFlow/Keras in a Docker environment. The model accepted $482\times482\times3$ RGB inputs represented as float32 values scaled to the interval $[0,1]$. A pretrained MobileNetV2 backbone with width multiplier $\alpha=0.35$ was truncated at the \texttt{block\_6\_expand\_relu} layer, producing a $61\times61$ spatial output grid. Batch-normalization momentum was set to 0.9. The detection head consisted of a $1\times1$ convolution with $32$ ReLU-activated channels followed by a $1\times1$ convolution producing logits for the implicit background class and the single explicit \texttt{turtle} class. A softmax layer was appended before model export.

The model training contained $10,215$ labeled frames derived from eight source footage sequences. Of these, $5,465$ frames contained one labeled turtle and $4,750$ contained no labeled turtle. Background only frames were retained as negative observations. Before final partitioning, the annotations were audited against the $482\times482$ pixel image boundaries. After correction of edge crossing annotations, the final audit identified no malformed or out of bounds bounding boxes.

Because adjacent frames from the same video can be nearly identical, individual frames were not assigned independently at random. Each footage source was divided into contiguous temporal blocks of approximately 60 frames. These blocks were partitioned into approximately $70\%$ training, $15\%$ validation, and $15\%$ held out test sets. Split construction balanced turtle positive frames, background frames, total bounding boxes, and small, medium, and large turtle instances across the three partitions. A five frame boundary guard parameter discouraged split transitions near turtle positive observations. The resulting training set contained $7,150$ frames, including $3,831$ turtle-positive and $3,319$ background frames. The validation set contained $1,531$ frames, including $852$ turtle-positive and $679$ background frames. The held out test set contained $1,534$ frames, including $782$ turtle positive and $752$ background frames. The held out partition was not used for checkpoint selection or for the validation results reported below.

Object detection augmentation was enabled during training. The configured augmentation pipeline included random resized crops to $482\times482$ pixels with scale factors from 0.90 to 1.10, rotations of up to $\pm20^\circ$, random brightness and contrast changes with limits of $0.20$ and probability $0.5$, and horizontal flips with probability $0.5$. Bounding box annotations were transformed together with the images and converted into spatial FOMO targets on the network's output grid.

Training used weighted cross entropy, with turtle positive output cells assigned a loss weight of 100 relative to background cells. The model was optimized with Adam using a learning rate of $10^{-3}$, a batch size of $32$, and a maximum of $120$ epochs. Validation precision, recall, and F1 were evaluated after every epoch using the Edge Impulse centroid scorer with a confidence threshold of $0.03$, touching-region fusion only \texttt{fuse\_gap\_cells}=0, and a normalized centroid-matching distance of $0.20$. The checkpoint with the highest validation F1 was retained and restored before export.

The selected checkpoint occurred at epoch $67$. After restoration and rescoring on the $1,531$ frame validation partition, the model achieved centroid assignment precision of $0.808994$, recall of $0.899462$, and F1 of $0.851833$, with a validation loss of $0.023697$. Rounded to three decimal places, these values correspond to precision $0.809$, recall $0.899$, and F1 $0.852$.

These validation statistics are specific to the Edge Impulse centroid assignment policy. They are not intersection over union measurements or mean average precision. FOMO predicts class activation regions rather than the full spatial extent of an object, and the scorer can credit multiple spatially distinct predictions lying within the accepted distance of one ground-truth turtle. The reported values should therefore be described as centroid-assignment precision, recall, and F1 rather than conventional bounding-box detection metrics.

The selected training run generated both a float32 TensorFlow Lite model and a full-integer TensorFlow Lite model with int8 input and output tensors. The corresponding model files were $82,492$ and $54,520$ bytes, respectively. A compiled Edge Impulse model was used for onboard execution on the Raspberry Pi 5.

During the onboard aquarium trial, the two forward facing camera streams were decoded and combined into a stitched wide field image. Edge Impulse's Studio compatible preprocessing prepared the stitched image for model inference.

Each camera frame was processed independently. From the turtle candidates returned by the Edge Impulse runtime, the perception node rejected candidates below an application-level confidence threshold of 0.5 and selected the highest-confidence remaining candidate. The candidate's image-plane centroid was transformed back into the coordinate system of the stitched image and published to the motion planner through the ROS $\texttt{centroids}$ topic. When no candidate was accepted, the node published an empty centroid message and recorded the tracking status as lost. The trained detector was deployed for live tracking of free-swimming sea turtles Myrtle and Carolina in the Giant Ocean Tank; full per-track results are reported in Table~\ref{tab:fomo_live}.

\section*{Supplementary Movie}
\noindent\textbf{Supplementary Movie 1.} Overview of the Crush platform and its
autonomous capabilities, showing notable trials in the Giant Ocean Tank at the
New England Aquarium: untethered obstacle avoidance, operator-initialized tracking of
arbitrary targets, and fully onboard autonomous
tracking of a free-swimming sea turtle. Except where noted, all perception and control run onboard.

\begin{table}[ht]
\centering
\caption{\textbf{Live untethered FOMO tracking trials (Giant Ocean Tank).}
Free-swimming sea turtles Myrtle and Carolina were tracked using the onboard
FOMO detector in fully untethered mode.}
\label{tab:fomo_live}
\begin{tabular}{@{}lrcp{5.2cm}@{}}
\toprule
Animal & Duration (s) & Failed & End track reason \\
\midrule
Green sea turtle (Myrtle)        & 70 & No  & Stopped swimming\textsuperscript{a} \\
Green sea turtle (Myrtle)        & 34 & Yes & Tracker failure --- missed detection \\
Green sea turtle (Myrtle)        & 16 & Yes & Tracker failure --- false positive (coral misdetected as turtle) \\
Green sea turtle (Myrtle)        & 17 & Yes & Specular/mirror-induced mistrack \\
Loggerhead sea turtle (Carolina) & 12 & No  & Coral collision\textsuperscript{b} \\
Loggerhead sea turtle (Carolina) & 13 & No  & Coral collision\textsuperscript{b} \\
\bottomrule
\end{tabular}
\vspace{0.3cm}
\begin{flushleft}
\footnotesize
\textsuperscript{a} Some tracks end early for the safety of the aquarium animals.\\
\textsuperscript{b} Track was not lost but ended early to prevent damage to the robot.
\end{flushleft}
\end{table}
\begin{table}[ht]
\centering
\caption{\textbf{Full analysis of operator-initialized pool and aquarium tracks.} Sea animals tracked in the Giant Ocean Tank at the New England Aquarium, and various objects tracked in the MIT Alumni Pool. Myrtle is a green sea turtle (\textit{Chelonia mydas}); Carolina is a loggerhead sea turtle (\textit{Caretta caretta}). Cownose rays are \textit{Rhinoptera bonasus}; great barracuda are \textit{Sphyraena barracuda}.}
\label{tab:track_full_analysis}
\begin{tabular}{@{}llcl@{}}
\toprule
Animal/Object Type & Duration (s) & Failed & End Track Reason \\
\midrule
Green sea turtle (Myrtle) & 59 & No & Stopped swimming \\
Green sea turtle (Myrtle) & 57 & No & Stopped swimming \\
Green sea turtle (Myrtle) & 43 & No & Visibility \\
Green sea turtle (Myrtle) & 38 & No & Coral collision\textsuperscript{b} \\
Green sea turtle (Myrtle) & 36 & No & Depth \& visibility \\
Loggerhead sea turtle (Carolina) & 29 & No & Depth limitation \\
Loggerhead sea turtle (Carolina) & 23 & Yes & Tracker failure \\
Green sea turtle (Myrtle) & 22 & No & Coral collision\textsuperscript{b} \\
Green sea turtle (Myrtle) & 20 & Yes & User error \\
Green sea turtle (Myrtle) & 12 & No & Swam into robot\textsuperscript{a} \\
Loggerhead sea turtle (Carolina) & 9 & No & Depth limitation \\
Loggerhead sea turtle (Carolina) & 8 & No & Depth limitation \\
Cownose ray & 32 & No & Swam over robot \\
Cownose ray & 30 & No & Too fast \\
Cownose ray & 22 & Yes & Tracker failure \\
Cownose ray & 15 & No & Swam behind robot \\
Cownose ray & 14 & Yes & Tracker failure \\
Cownose ray & 13 & No & Swam under \\
Great Barracuda & 20 & No & Depth limitation \\
Great Barracuda & 15 & Yes & Tracker failure \\
Diver & 45 & No & --- \\
Diver & 29 & No & --- \\
Toy Turtle & 249 & No & --- \\
Toy Turtle & 245 & No & --- \\
Toy Turtle & 223 & No & --- \\
Toy Turtle & 169 & No & --- \\
Toy Turtle & 161 & No & --- \\
Toy Turtle & 116 & No & --- \\
Toy Turtle & 96 & No & --- \\
Toy Turtle & 96 & No & --- \\
Toy Turtle & 83 & No & --- \\
Toy Turtle & 79 & No & --- \\
Toy Turtle & 71 & No & --- \\
Toy Turtle & 70 & No & --- \\
Toy Turtle & 57 & No & --- \\
Toy Turtle & 50 & No & --- \\
Toy Turtle & 47 & No & --- \\
Toy Turtle & 45 & No & --- \\
Toy Turtle & 36 & No & --- \\
Toy Turtle & 28 & No & --- \\
Toy Turtle & 23 & No & --- \\
Toy Turtle & 22 & No & --- \\
Toy Turtle & 20 & No & --- \\
Toy Pig & 133 & No & --- \\
Toy Pig & 67 & No & --- \\
Toy Pig & 21 & No & --- \\
Fish Tail & 106 & No & --- \\
Human & 93 & No & --- \\
Human & 79 & No & --- \\
Human & 65 & No & --- \\
Human & 31 & No & --- \\
Plastic Head & 92 & No & --- \\
\bottomrule
\end{tabular}

\vspace{0.3cm}
\begin{flushleft}
\footnotesize
\textsuperscript{a} Some tracks end early for the safety of the aquarium animals.

\textsuperscript{b} Track was not lost but ended early to prevent damage to the robot.
\end{flushleft}
\end{table}

\begin{table}[ht]
\centering
\footnotesize
\setlength{\tabcolsep}{4pt}
\caption{\textbf{Comparison of Crush with representative biomimetic underwater
robots.} Platforms are compared on physical performance and on the level of
autonomy demonstrated.}
\label{tab:comparison}
\begin{tabular}{@{}lcccccccc p{2.4cm}@{}}
\toprule
& \multicolumn{3}{c}{Physical performance} & \multicolumn{5}{c}{Autonomy and deployment} & \\
\cmidrule(lr){2-4} \cmidrule(lr){5-9}
Platform & Mass & Speed & CoT\textsuperscript{a}
         & Tether\textsuperscript{b} & Depth & Onboard & Onboard & Animal & \\
         & (kg) & (m\,s\textsuperscript{-1}) & (---)
         &  & control & perception & power & tracking
         & \raisebox{1.6ex}{Test 
         environment} \\
\midrule
\textbf{Crush}
  & 10   & 0.208 & 2.9  & Both & Yes & Yes & Yes & Yes & Pool, tank, aquarium, ocean \\
ART~\cite{baines2022multi}
  & 9    & 0.083 & 3.0  & Yes  & No  & No  & No  & No  & Pool, lake \\
Van der Geest turtle~\cite{vandergeest2023employing}
  & 2.5  & 0.48  & 3.0  & Yes  & No  & No  & No  & No  & Tank \\
Tunabot Flex~\cite{white2021tunabotflex}
  & 0.19 & 1.17  & 1.88 & Yes  & No  & No  & No  & No  & Flow tank \\
SoFi~\cite{katzschmann2018exploration}
  & 1.6  & 0.235 & 14   & No   & Yes & Yes & Yes & No  & Pool, ocean \\
Deep-sea soft robot~\cite{li2021self}
  & ---\textsuperscript{c} & 0.052 & ---\textsuperscript{c}
  & No & No & No & Yes & No & Chamber, lake, ocean \\
\bottomrule
\end{tabular}
\vspace{0.3cm}
\begin{flushleft}
\footnotesize
\textsuperscript{a} Dimensionless cost of transport at each platform's maximum
reported speed, $\mathrm{CoT}=P/(mgU)$. Values for other platforms are
computed from figures reported in the cited work. Cross-platform comparison is
approximate because the system boundary on $P$ differs between studies.

\textsuperscript{b} ``Both'' indicates the platform has been operated in both
tethered and untethered configurations.

\textsuperscript{c} Mass is not reported, so cost of transport cannot be
computed. The onboard high-voltage amplifier draws approximately 3.8\,W,
giving a mass-independent work per meter of roughly
73\,J\,m\textsuperscript{-1} at the reported swimming speed.
\end{flushleft}
\end{table}

\end{document}